\documentclass{article}
\usepackage{graphicx}
\usepackage{subfigure}
\usepackage{float}
\usepackage{hyperref}

\begin{document}
\title{Open data for Moroccan license plates for OCR applications : data collection, labeling, and model construction.}
%
%
\author{Abdelkrim Alahyane $^1$ \and
Mohamed El Fakir $^1$ \and
Saad Benjelloun $^1$ \and 
Ikram Chairi $^1$}

\date{ $^1$ MSDA group, Mohammed VI Polytechnic University (UM6P), Lot 660, Hay Moulay Rachid, Ben Guerir 43150 , Morocco}
%
%

%
\maketitle              
\begin{abstract}
Significant number of researches have been developed recently around intelligent system for traffic management, especially, OCR based license plate recognition, as it is considered as a main step for any automatic traffic management system. Good quality data sets are increasingly needed and produced by the research community to improve the performance of those algorithms. Furthermore, a special need of data is noted for countries having special characters on their licence plates, like Morocco, where Arabic Alphabet is used. 
In this work, we present a labeled open data set of circulation plates taken in Morocco, for different type of vehicles, namely cars, trucks and motorcycles. This data was collected manually and consists of 705 unique and different images.  Furthermore this data was labeled for plate segmentation and for matriculation number OCR. Also, As we show in this paper, the data can be enriched using data augmentation techniques to create training sets with few thousands of images for different machine leaning and AI applications. We present and compare a set of models built on this data. Also, we publish this data as an open access data to encourage innovation and applications in the field of OCR and image processing for traffic control and other applications for transportation and heterogeneous vehicle management. 

\end{abstract}
\section{introduction}
The world is knowing a continuously increase of road traffic since early 2000s. Traffic management is one of the main solutions to avoid congestion and road accident. With the recent technological evolution concerning IoT and smart sensors, many researches have addressed road traffic issues using data generated by transportation and mobility systems\cite{REF1}.

One of the most developed sub-field of smart mobility, is Automatic Number Plate Recognition (ANPR), which is an integration of Artificial Intelligence along with computer vision and pattern recognition \cite{REF2,REF3}. ANPR consists of three main stages: Number Plate Localization, character segmentation and Optical character recognition (OCR). 

OCR  plays a main role in ANPR as it transforms character into encoded text information. Different methods for OCR were proposed in the literature going from Thresholding \cite{REF5} to Machine Learning methods \cite{REF3}. 
To improve performance of recognition systems, especially for Machine Learning methods, algorithms need to be trained on a big amount of data representing real situation of cars in road traffic. For this reason, different platforms provide access to data set for car plate images like Kaggle, Github or different open data web site. However, as each country has its own way to characterize  plate numbers, those data bases can not be used for all countries, especially the ones using Arabic alphabet like Morocco. Up to today, there is no data set containing Moroccan vehicle plates in open data platforms. 

In this work, we present a first Moroccan vehicle plate data set containing 705 unique labeled images for cars, motocycles and trucks. The data set was collected by taking randomly pictures for vehicles in different context.
We present in this article also different way to manipulate this data set and to possibility use data augmentation techniques to improve learning of Machine Learning recognition algorithms. We present also baseline models for plate segmentation and recognition using the built data set.
The rest of the paper is organized as follows. In section 2, we present the data set, in section 3 we focus on describing the labeling step and in section 4 we present baseline models for the plate segmentation and OCR tasks. We conclude the article in section 5. 

\section{Presentation of the data set}
The natural first step in building plate segmentation and recognition models is to collect an adequate data-set bringing together vehicle's images with relatively clear licence plates taken in different contexts and environments. The goal is to capture all possible variations related to the surroundings of real world plates and thus render the built AI models more robust and efficient.

The set of images collected must at least respond to two major practical issues. On the one hand, it must allow the construction of models ensuring the segmentation of license plates in images of vehicles. On the other hand, it is supposed to grant the detection and automatic recognition of the characters constituting the plates by training suitable OCR models.

\subsection{Data for license plate segmentation}

\begin{figure}[H]
\includegraphics[width=\textwidth]{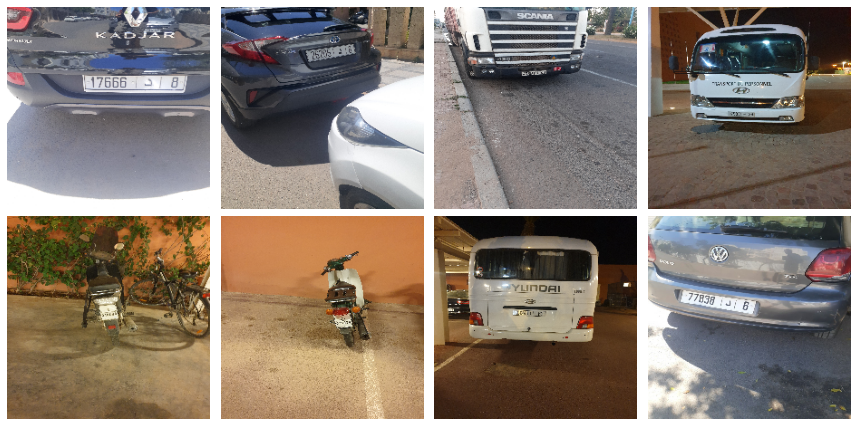}
\caption{Sample of 8 images taken from the original data-set} \label{fig:plates}
\end{figure}

The performance of models ensuring the plate segmentation task depends deeply on the quantity and quality of the collected images used for training. As mentioned above, multiple environments (views, backgrounds, scenes...) must be taken into consideration when collecting the data to ensure that models can properly generalize their learning. Therefore, we were interested in collecting and building a set of good quality images with different contexts.
To achieve this, the data was gathered by randomly capturing pictures of vehicles in Moroccan streets, the resulting data-set contains 705 unique images of different types of vehicles (motorcycles, cars and trucks) with two main categories of Moroccan license plates: square and rectangular plates. Some examples of those different vehicle types are shown in figure \ref{fig:plates}, while the number of images corresponding to each type of vehicle are specified in table \ref{tab:plates}. Another important point to mention is that some of the images contain multiple licence plates.

\begin{table}
\caption{Number of images aggregated by vehicle type.}\label{tab:plates}
\begin{center}
\label{tab:vehicule_types}
\begin{tabular}{|c|c|c|c|}
\hline   Vehicles types & Cars  & motorcycles  & Trucks \\
\hline  Number of images & 633 & 35 & 37  \\
\hline 
\end{tabular}
\end{center}
\end{table}

The shooting angles of the images as well as the zoom degrees are also multiple and very diversified. This always aims to capture as much as possible, the variability of conditions in which the trained model could operate and predict. 

As it will be detailed later, this data set was labeled in order to be used to train machine learning algorithm for plates segmentation.

Finally we note that the data can be enriched using data augmentation techniques such as adding noise, changing contrast, flipping and rotating, to create training sets with four to five thousands (4000 to 5000) images, to train robust machine leaning and AI applications.

\subsection{Data for plates OCR}

After collecting this data-set, rich enough in images of vehicles that can be used to train a plate segmentation model, we cropped each image to extract only the license plates. This aims to build a data-set of license plate images which will be subsequently annotated to train OCR models allowing the automatic segmentation and recognition of characters.

\begin{figure}[H]

\begin{subfigure}{}
\begin{center}
\includegraphics[width=0.7\textwidth]{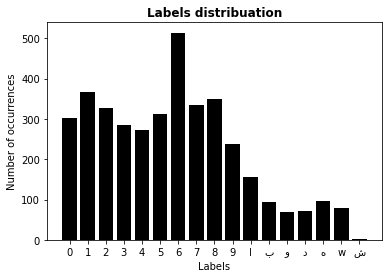}
\caption{Labels distribution in the plate data-set.}
\label{fig:chars}
\end{center}
\end{subfigure}

\begin{subfigure}{}
\begin{center}
\includegraphics[width=0.5\textwidth]{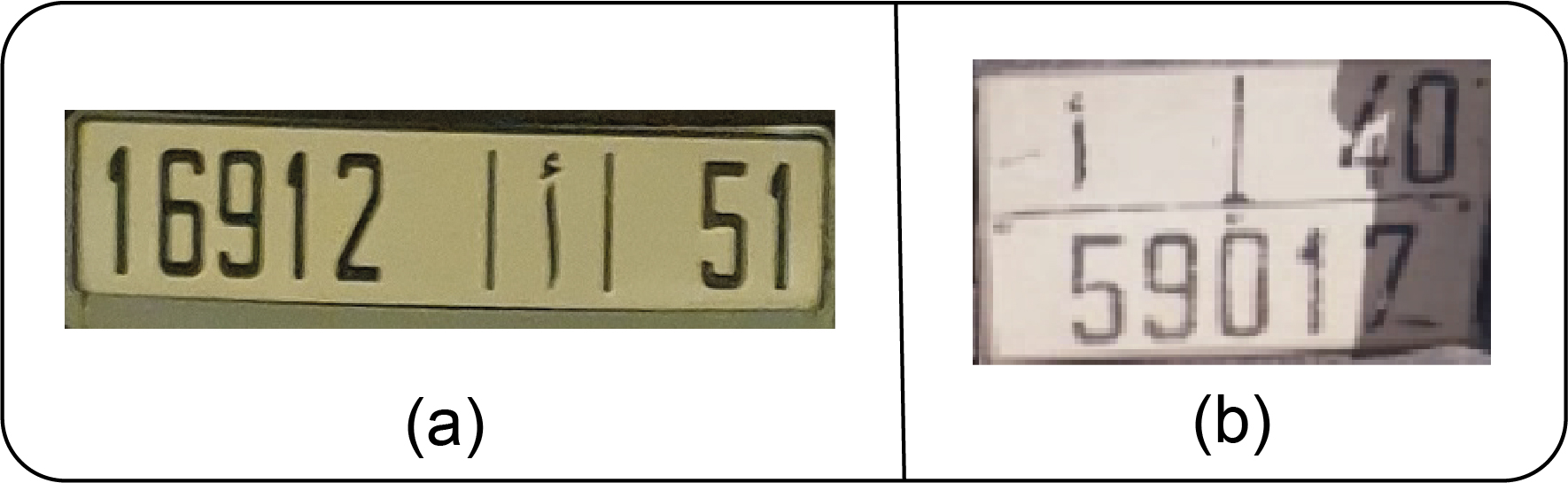}
\caption{License plate types.} \label{fig:forms}
\end{center}
\end{subfigure}
\end{figure}

The plates obtained contains a set of 17 characters, 10 digits (from 0 to 9), and 7 Arabic letters corresponding to the letters currently used in Moroccan license plates. The distribution of these different labels is shown in the histogram of figure \ref{fig:chars}. It should also be noted, as mentioned previously, that the plates are in two main categories: the rectangular plates represented by the plate (a) of figure \ref{fig:forms}, then the square plates, generally used for motorcycles and some trucks, represented by the plate (b) of the same figure.

\subsection{Data preparation}

After gathering the data-sets, the next step is to prepare the images to be used to solve the two problems mentioned previously. Indeed, the construction of high-performance models requires a significant amount of data, that's why data augmentation can be a good solution. This is a technique allowing a significant increase in the quantity of images available to us. It is based on the principle of generating new observations or images starting from the existing ones. In our case, we performed the following data augmentation operations:
\begin{itemize}
  \item Flipping the image vertically and horizontally.
  \item Shearing, which means shifts part of the image like a parallelogram.
  \item Adding noise, we choose to use a noise sampled from Gaussian distributions.
  \item Adjust images brightness.
  \item Apply affine transformations and scaling.
\end{itemize}
Data augmentation has demonstrated its effectiveness especially on the training of neural networks which requires a significant amount of images. After completing this step we obtained 4935 images.

\paragraph{}We limit ourselves to these preparation operations and potential future users of these data-sets can make all the appropriate pre-processing steps according to the models that will be built and their respective architectures. Other suggestions for effective preparation steps are as follows: Re-perform the data augmentation operations to further increase not only the quantity of images used for learning but also to add noise in order to improve generalization, especially if the models used are neural networks and if the computing power is encouraging. Some other approaches and algorithms can be more precise by filtering the input images using noise reduction and smoothing filters, for others it may be necessary to remove the background color from the images to reduce the noise. If one wants to reduce the number of parameters to estimate, thereby reducing the complexity of the model, it may be more interesting to perform  gray scaling operations or binarization.

In conclusion, the choice of data preparation operations depends on many factors, including the nature of the models, their characteristics and the available computing power. 

\section{Data Labeling}

 
Labeling data is a major step in a supervised machine learning tasks, as the model's output depends on the labels we introduce during the learning phase. The quality of the training data determines the quality of the model, the same is true for the quality of annotation and its correctness. 

There are several types of annotations for image segmentation. In our case we use Bounding Boxes. Using an annotation tool we draw rectangular boxes around the plates to segment them in the vehicle images. The traced boxes are then represented by specific files containing the coordinates of the x and y axes in the upper left corner and the coordinates of the x and y axes in the lower right corner of the rectangle.

The data set we present is labeled in three formats, to give researchers and users the freedom to use multiple models that require different annotation formats. These formats are as following:

\subsection{XML format}
\begin{figure}[H]
\begin{center}
\includegraphics[width=0.5\textwidth]{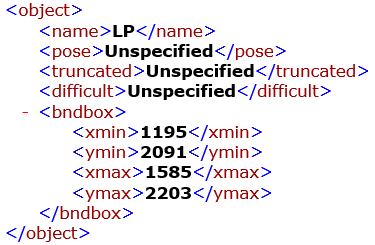}
\caption{The object tag of the XML file, it contains the labeling information specific to a Bounding Box, including coordinates and label name.} \label{fig:xml}
\end{center}
\end{figure}

For each vehicle image, we generate an XML file to annotate the plates it contains. The same process will be performed for the images of the plates used for the OCR in order to segment the characters.
This format allows us to transfer a lot of useful information about the bounding boxes in the images namely: Xmin,Ymin (upper left corner coordinates) and Xmax, Ymax (lower right corner coordinates). The XML format also contains the name of the class corresponding to each traced box. Furthermore, we can evaluate on a scale of 0 to 1 whether the object is difficult or truncated. For reasons of simplicity we have chosen not to specify these last two parameters. An example of XML content describing the annotation information for a single bounding box in the image is shown in Figure \ref{fig:xml}.

\subsection{Text file format}
\begin{figure}[H]
\begin{center}
\includegraphics[width=160pt,height=75pt]{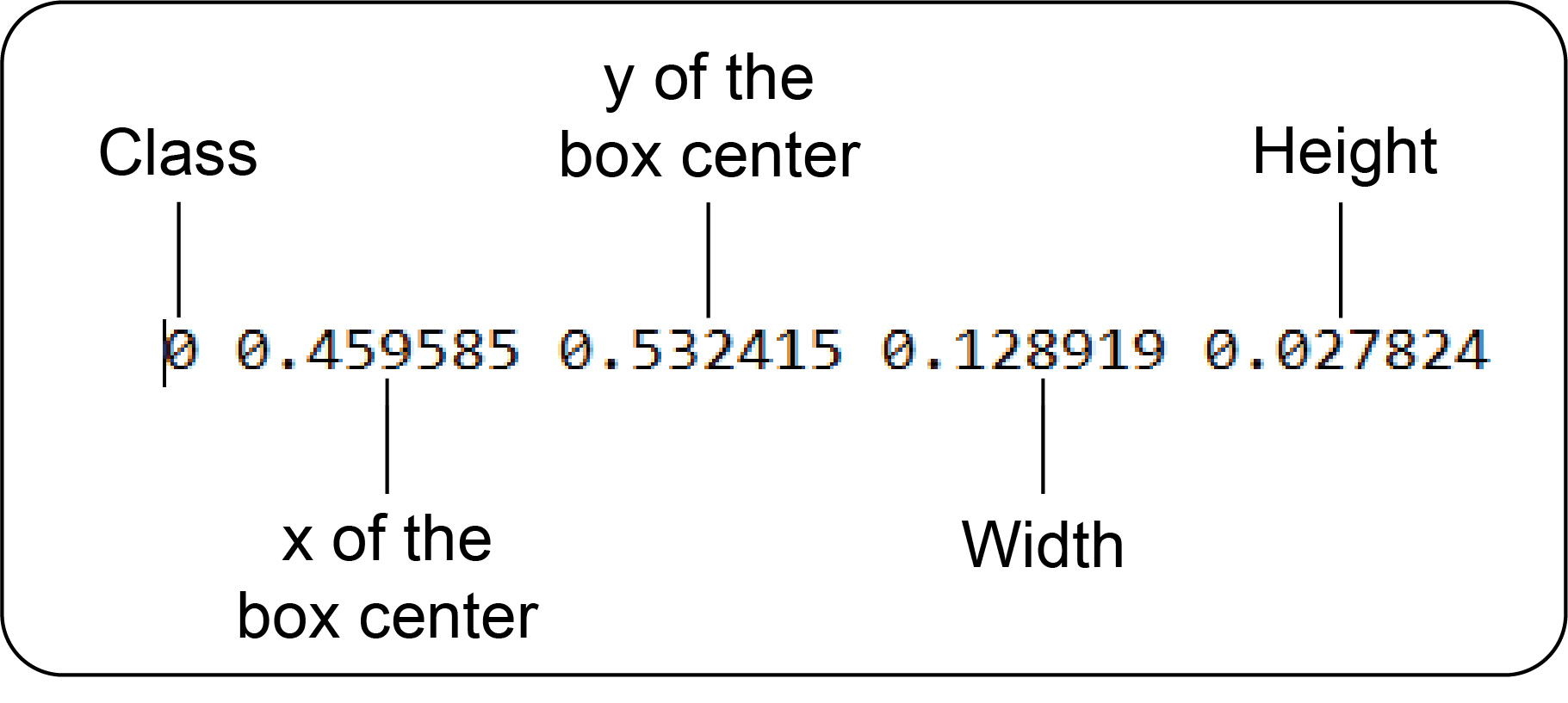}
\caption{Line of annotation text file with clarifications concerning the various information it contains.} \label{fig:txt}
\end{center}
\end{figure}

The second annotation format is a Text file adapted to YOLO \cite{REF8} requirements (normalized coordinates).
This format is very simple, again we generate a text file for each image, each line contains an object labeled in the image by carrying the coordinates of the center of its boundary box, the height and the width of the rectangular box and the detected class. An example of a line from the text file representing an annotated object in the image is shown in figure \ref{fig:txt}.

\subsection{CSV format}
\begin{figure}[H]
\begin{center}
\includegraphics[width=\textwidth]{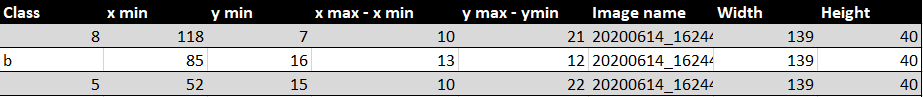}
\caption{Three (3) lines of the CSV file used to annotate the data-set.} \label{fig:csv}
\end{center}
\end{figure}

The third format that we make available to researchers is the CSV format, which carries like that of XML: the coordinates, the height and width of the delimiting box, the class of the object and finally the name of the image. The only exception is that here is a single CSV file concerning all the images, with each line representing information concerning a single bounding box. An example of the contents of the lines of the csv file is shown in figure \ref{fig:csv}.

\section{Baseline models} 
The objective of this section is to present some models showing how one can use the data set to build powerful AI applications. These models can also serve as baseline to compare their performance with future models. Therefore, we will build some machine and deep learning applications trying to solve the two problems mentioned above, namely Plate segmentation and plate recognition using OCR models.

\subsection{Plate segmentation}
Given its performance, speed and precision, the YOLOv3 \cite{REF8} architecture is the model chosen to be developed and trained on plate segmentation, we will use a training-set of 650 images and a testing-test of 55 images. This set corresponds to original images without data augmentation. Another baseline model to consider is a pre-trained model on plate segmentation, which is the Wpod-Net model \cite{REF7}. It is a convolutional neural network model built from a data-set containing images of foreign (i.e. non Moroccan) vehicle licence plates. These vehicle images correspond to 10 different countries: Germany, Vietnam, Japan, Thailand, Saudi Arabia, Russia, North Korea, USA, India and China. The goal is to evaluate this model on our testing-set containing vehicles with Moroccan licence plates, and use it as a baseline model to compare these performances with those of YOLO v3.
\paragraph{}We will use the accuracy metric on the testing-set as a performance measure to identify the proportion of plates detected or segmented by each of the two models. The results, as presented in table \ref{tab:seg}, show that the YOLO v3 architecture performs better than the Wpod-Net model despite the small number of images that we used to train the YOLO neural network. Besides that, even though the Wpod-Net is trained using a gigantic amount of vehicle images from several countries and it has performed well in many experiments, it could not outperform the YOLO model. This is explained by the fact that the Wpod-Net model has never seen vehicle images of Moroccan license plates. So although the plates have a lot of common characteristics, it was not able to detect all of them. Unlike the YOLO model which, despite the small training set used, succeeded in segmenting the overwhelming majority of the plates present in the test-set.

\begin{table}
\begin{center}
\caption{Results of used models.}\label{tab:seg}
\label{tab:vehicule_types}
\begin{tabular}{|c|c|c|}
\hline  Models & Accuracy Training set  & Accuracy Testing set   \\
\hline  Wpod-Net & (pre-trained) & 0,7077   \\
\hline  YOLOv3 & 0,9861 & 0,8506   \\
\hline 
\end{tabular}
\end{center}
\end{table}

\paragraph{}This illustrates the importance of the data-set that we have grouped in the Moroccan plate segmentation applications. It can be used, for example, to perform transfer learning and improve already existing models that allow segmentation of foreign plates.

\subsection{Plate OCR}
Concerning the recognition of the plates, we will use three different approaches namely:
\begin{itemize}
  \item Histogram projection with k-NN classifier.
  \item Histogram projection with CNN classifier.
  \item YOLO v3, to detect characters in the plate and classify them simultaneously.
\end{itemize}
The thresholds of the histogram projection as well as the hyper-parameter $k$ are fixed by using the cross validation approach. On the other hand, the CNN architecture used is MobileNetV2 by Google \cite{REF6}. The training-set is composed of characters from 650 plate images while the models are tested using characters from 55 plate images. The results concerning the accuracy are shown in table \ref{tab:ocr}: 
\begin{table}
\begin{center}
\caption{Results of OCR used models.}\label{tab:ocr}
\label{tab:vehicule_types}
\begin{tabular}{|c|c|c|}
\hline    Models & Accuracy Training set  & Accuracy Testing set   \\
\hline  Histogram projection + k-NN classifier & 0,5128 & 0,2187   \\
\hline  Histogram projection + CNN classifier & 0,6894 & 0,3775   \\
\hline  YoloV3 & 0,9142 & 0,8238   \\
\hline 
\end{tabular}
\end{center}
\end{table}

We can therefore observe that the YOLO model again has the best performance. It will then be a possible baseline architecture to compare with other models.

\section{Conclusion}

In this work we presented a Labeled data set of multiple-type vehicle images, taken in Morocco. We presented the process of labeling this data for plate segmentation and for plate's number OCR using different formats. We put this labeled data openly available for the research, startup and industrial community to build models for applications related to traffic management or car parks management in Morocco. We believe that initiatives such as ours can catalyse the innovation and the technological development of AI solutions in countries like Morocco by removing the burden linked to the non-availability of open data and to the time-consuming tasks of collecting and labeling data. 
We presented a set of Machine learning models that can be used as baseline models and to which future users of this data set can compare and aim to outperform by innovating in term of computational methods and algorithms.
The labeled data set, with data augmentation or without, can be downloaded at \cite{msda_datasets}, and all the implemented algorithms are available at \cite{algorithms}.

\end{document}